\documentclass[pmlr]{jmlr}%

\usepackage{longtable}%

\usepackage{booktabs}
\usepackage{graphicx}

\usepackage{cleveref}
\usepackage{alltt}
\usepackage[most]{tcolorbox}
\usepackage{tikz}

\tcbset{
  aibox/.style={
    width=\textwidth,
    top=10pt,
    colback=white,
    colframe=black,
    colbacktitle=black,
    enhanced,
    center,
    attach boxed title to top left={yshift=-0.1in,xshift=0.15in},
    boxed title style={boxrule=0pt,colframe=white,},
  }
}
\newtcolorbox{AIbox}[2][]{aibox,title=#2,#1}

\definecolor{aigold}{RGB}{244,210, 1} 
\definecolor{aired}{RGB}{255,180,181}

\makeatletter
\def\set@curr@file#1{\def\@curr@file{#1}} %
\makeatother

\newcommand{\eat}[1]{\ignorespaces}

\theorembodyfont{\upshape}
\theoremheaderfont{\scshape}
\theorempostheader{:}
\theoremsep{\newline}

\jmlrvolume{219}
\jmlryear{2023}
\jmlrworkshop{Machine Learning for Healthcare}

\usepackage[nolist]{acronym}

\makeatletter
\newcommand*{\org@overidelabel}{}
\let\org@overridelabel\@verridelabel
\@ifpackagelater{acronym}{2015/03/21}{%
  \renewcommand*{\@verridelabel}[1]{%
    \@bsphack
    \protected@write\@auxout{}{\string\AC@undonewlabel{#1@cref}}%
    \org@overridelabel{#1}%
    \@esphack
  }%
}{%
  \renewcommand*{\@verridelabel}[1]{%
    \@bsphack
    \protected@write\@auxout{}{\string\undonewlabel{#1@cref}}%
    \org@overridelabel{#1}%
    \@esphack
  }%
}
\makeatother

\title[Scaling Clinical Trial Matching using LLMs]{Scaling Clinical Trial Matching Using Large Language Models: A Case Study in Oncology}

\author{\Name{Cliff Wong}
       \Email{cliff.wong@microsoft.com}\\ 
       \addr Microsoft Research\\
       \addr Redmond, WA, USA
       \AND
       \Name{Sheng Zhang}
       \Email{zhang.sheng@microsoft.com}\\ 
       \addr Microsoft Research\\
       \addr Redmond, WA, USA
       \AND
       \Name{Yu Gu}
       \Email{aiden.gu@microsoft.com}\\ 
       \addr Microsoft Research\\
       \addr Redmond, WA, USA
       \AND
       \Name{Christine Moung}
       \Email{christine.moung@providence.org}\\ 
       \addr Providence Health \& Services Molecular Genomics Laboratory\\
       \addr Portland, OR, USA
       \AND
       \Name{Jacob Abel}
       \Email{jacob.abel@providence.org}\\
       \addr Providence Health \& Services Molecular Genomics Laboratory\\
       \addr Portland, OR, USA
       \AND
       \Name{Naoto Usuyama}
       \Email{naotous@microsoft.com}\\
       \addr Microsoft Research\\
       \addr Redmond, WA, USA
       \AND
       \Name{Roshanthi Weerasinghe}
       \Email{roshanthi.weerasinghe@providence.org}\\ 
       \addr Clinical Research Analytics, Providence Health \& Services\\
       \addr Portland, OR, USA
       \AND
       \Name{Brian Piening}
       \Email{brian.piening@providence.org}\\
       \addr Earle A. Chiles Research Institute, Providence Cancer Institute\\
       \addr Portland, OR, USA
       \AND
       \Name{Tristan Naumann}
       \Email{tristan@microsoft.com}\\ 
       \addr Microsoft Research\\
       \addr Redmond, WA, USA
       \AND
       \Name{Carlo Bifulco}
       \Email{carlo.bifulco@providence.org}\\ 
       \addr Earle A. Chiles Research Institute, Providence Cancer Institute\\
       \addr Portland, OR, USA
       \AND
       \Name{Hoifung Poon}
       \Email{hoifung@microsoft.com}\\ 
       \addr Microsoft Research\\
       \addr Redmond, WA, USA
       }

\begin{document}

\maketitle
\begin{acronym}
    \acro{DNF}{disjunctive normal form}
    \acro{EMR}{electronic medical record}
    \acro{IDN}{integrated delivery network}
    \acro{IRB}{institutional review board}
    \acro{LLM}{large language model}
    \acro{NLP}{natural language processing}
    \acro{PHI}{protected health information}
    \acro{U.S.}{United States}
\end{acronym}

\begin{sloppypar}
\clearpage
\begin{abstract}
  Clinical trial matching is a key process in health delivery and discovery. In practice, it is plagued by overwhelming unstructured data and unscalable manual processing.
  In this paper, we conduct a systematic study on scaling clinical trial matching using large language models (LLMs), with oncology as the focus area. 
  Our study is grounded in a clinical trial matching system currently in test deployment at a large U.S. health network.
  Initial findings are promising: out of box, cutting-edge LLMs, such as GPT-4, can already structure elaborate eligibility criteria of clinical trials and extract complex matching logic (e.g., nested AND/OR/NOT). While still far from perfect, LLMs substantially outperform prior strong baselines and may serve as a preliminary solution to help triage patient-trial candidates with humans in the loop.  
  Our study also reveals a few significant growth areas for applying LLMs to end-to-end clinical trial matching, such as context limitation and accuracy, especially in structuring patient information from longitudinal medical records.  
\end{abstract}

\section{Introduction}

Clinical trial matching identifies eligible patients to be considered for clinical trial enrollment, which is essential for clinical discovery and drug development. For diseases such as cancers where standard of care often fails, clinical trials are also a last hope and represent an important treatment option.
The standard process for clinical trial matching, however, is extremely inefficient.
Most information needed for accurate matching is scattered in vast amount of unstructured text, such as elaborate eligibility criteria of a clinical trial or patient information in longitudinal electronic medical records.
Curation and matching are currently done predominantly by manual effort, which is difficult to scale.
For example, in the \ac{U.S.} alone, there are two million new cancer patients every year and, at any given moment, there may be hundreds of active oncology trials~\citep{Siegel2022,clinicaltrials}. 
Manually evaluating all oncology trial-patient candidates is infeasible. 
Consequently, less than 3\% of patients were able to participate in a trial~\citep{Unger2019-ti}, whereas 38\% of trial failure stemmed from insufficient recruitment~\citep{stensland2014}. 

\Ac{NLP} has emerged as a promising direction for accelerating clinical trial matching. Prior work explores rule-based methods and supervised machine learning for structuring trial eligibility criteria~\citep{10.1136/amiajnl-2011-000321,10.1093/jamia/ocx019,yuan2019criteria,nye2020trialstreamer} or matching clinical trials against structured patient records~\citep{zhang2020deepenroll,gao2020compose}.
While promising, these methods still require extensive manual effort for rule development and example annotation. 
Recently, \acp{LLM}, such as GPT-4, have demonstrated impressive zero-shot and few-shot capabilities in both general domains~\citep{openai2023gpt4,bubeck2023sparks} and health applications~\citep{doi:10.1056/NEJMsr2214184,lee2023ai,nori2023capabilities}. 
In this paper, we conduct a systematic study on scaling clinical trial matching with \acp{LLM} by leveraging emergent in-context learning capabilities~\citep{NEURIPS2020_1457c0d6}.

\begin{figure*}[!t]
    \centering
    \includegraphics[width=\textwidth]{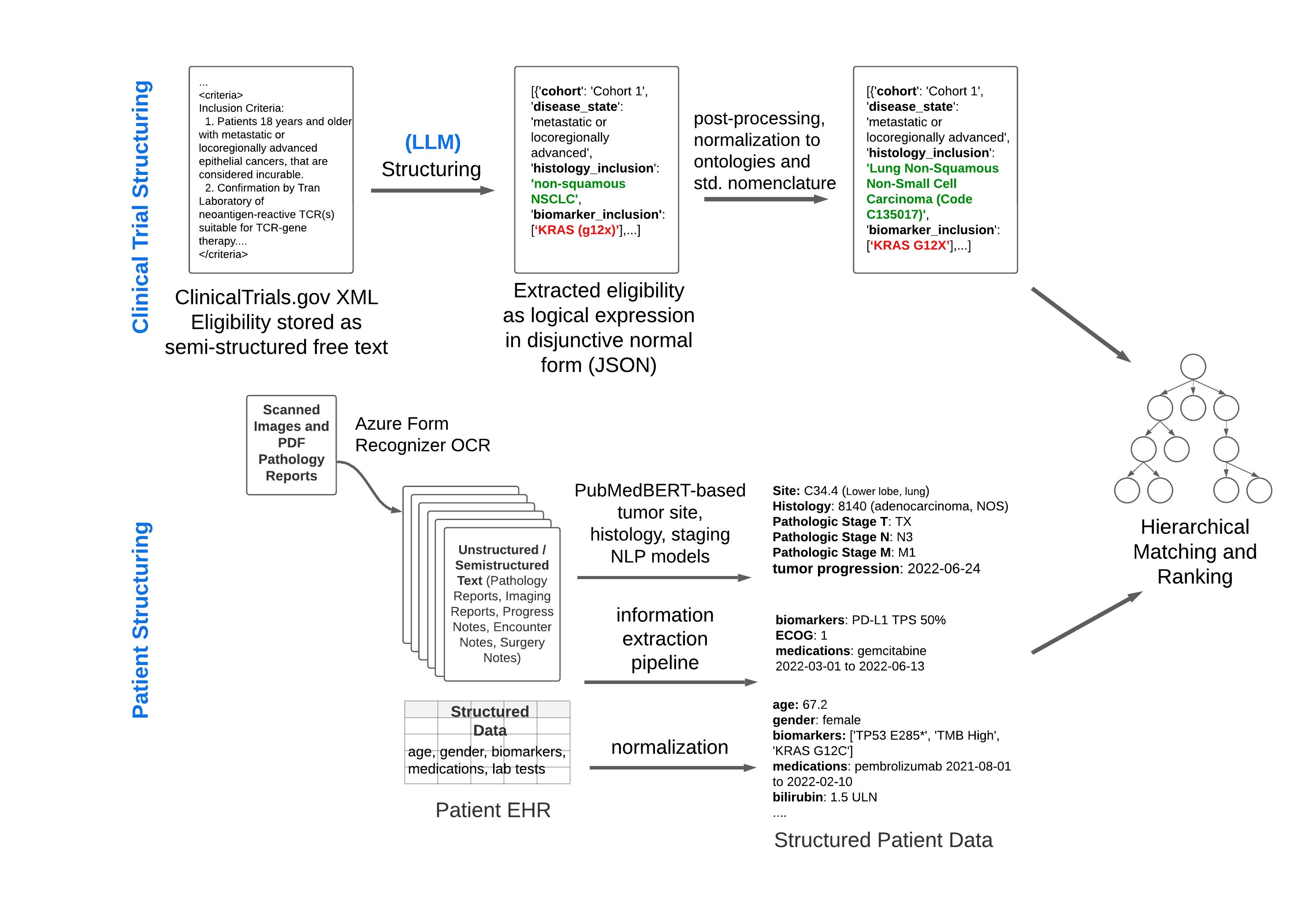}
    \caption{Overall schematic of matching. 
    }
    \label{fig:matching_diagram}
\end{figure*}

We choose oncology as the focus of our study because it incurs a high death burden and represents a large proportion of clinical trials (over 20\% in 2022\footnote{As of April 2023, \url{ClinicalTrials.gov} has 32,341 trials with start date 01/01/2022--12/31/2022; further filtering for `cancer' as disease yields 6,704 results. This reflects general trend with 449,665 available without start date restriction, of which 96,481 results include `cancer'.}).
Oncology trials also tend to contain elaborate eligibility criteria, such as complex combination logic of disease histology and genomic biomarkers, with the latter (genomic biomarkers) typically out of scope for prior state of the art~\citep{yuan2019criteria,zhang2020deepenroll,gao2020compose}.
We grounded our study in a real-world clinical trial matching system currently in test deployment at a large \ac{U.S.} health \ac{IDN}, which comprises three key components: structuring clinical trial eligibility criteria, structuring patient information from electronic medical records, and matching (\Cref{fig:matching_diagram}).
The patient structuring component leverages state-of-the-art self-supervised deep learning~\citep{preston2023cellpattern}. The trial-structuring and matching components are rule-based expert systems that require over 450 expert hours to develop.

For consistency and direct comparison with prior methods, we focus our study of \acp{LLM} primarily on structuring trial eligibility, but we also conduct a preliminary exploration on applying \acp{LLM} to end-to-end clinical trial matching.  

Initial results are promising: using merely up to three examples, state-of-the-art \acp{LLM}, such as GPT-4, can already structure elaborate trial eligibility criteria and extract complex matching logic of disease histology and genomic biomarkers. 
We conduct both intrinsic evaluation on trial structuring and end-to-end matching evaluation, using expert-annotated test sets and legacy enrollment data, respectively\footnote{Our code is available at \url{https://aka.ms/ctm-llm}}. 
While still far from perfect, \acp{LLM} substantially outperform prior strong baselines such as Criteria2Query~\citep{yuan2019criteria} and demonstrate competitiveness even against the oncology-specific expert system that requires many expert hours to develop and tailor for this domain.

\begin{figure}[!t]
    \centering
    \includegraphics[width=0.98\textwidth]{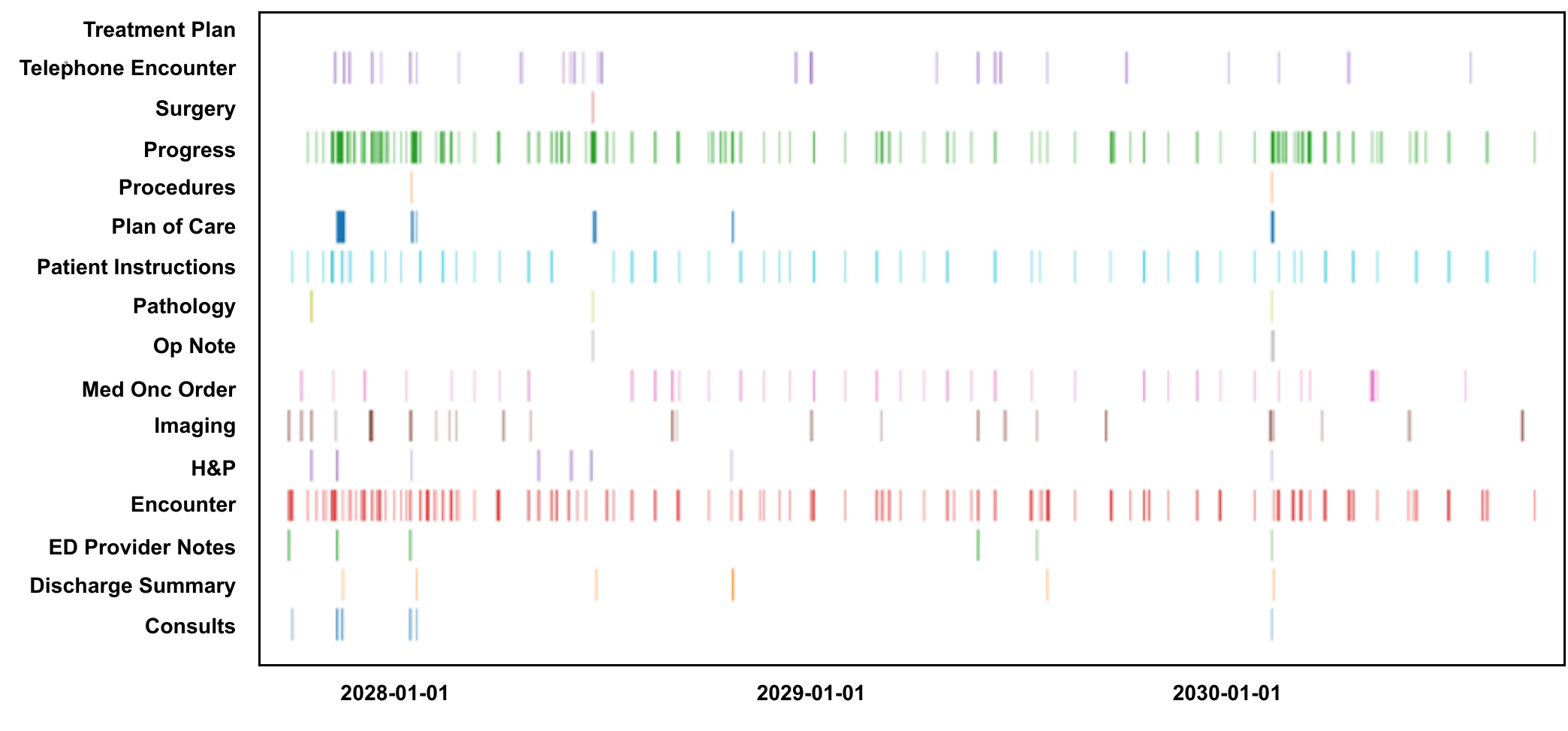}
    \caption{Patient de-identified timeline with various notes.}
    \label{fig:patient-journey}
\end{figure}

Our study also reveals significant growth areas for applying \acp{LLM} to end-to-end clinical trial matching, especially in structuring patient information from electronic medical records.
A cancer patient may have hundreds of notes, with key information such as tumor histology and biomarkers scattered across multiple notes~\citep{preston2023cellpattern} as shown in \Cref{fig:patient-journey}. Naively concatenating all potentially relevant notes will almost always exceed even the largest context size available for GPT-4: 32K tokens.\footnote{While GPT-4 has a context length of 8,192 tokens, there is limited access to GPT-4-32K which has a context length of 32,768–context (about 50 pages of text): \url{https://openai.com/research/gpt-4}.}
It might also risk overwhelming the \ac{LLM} with too much irrelevant information. In this paper, we thus resort to using structured patient information extracted by the state-of-the-art self-supervised deep-learning systems~\citep{preston2023cellpattern} and evaluate the \ac{LLM}'s capabilities in matching such information against the trial eligibility criteria. Preliminary results are promising and we leave more in-depth exploration to future work.

\subsection*{Generalizable Insights about Machine Learning in the Context of Healthcare}
\begin{itemize}
    \item We present the first systematic study on using \acp{LLM} to scale clinical trial matching. While we demonstrate this approach on a case study in oncology, our findings motivate the exploration of this approach across other areas.
    \item Our findings indicate that out of box, state-of-the-art LLMs such as GPT-4 can already handle complex eligibility criteria of clinical trials and extract matching logic.
    \item We also identify several key growth areas for enhancing LLMs in end-to-end clinical trial matching, such as addressing the context limitation and accuracy issues, especially in extracting and structuring patient data from electronic medical records.
\end{itemize}

\section{Related Work}

The existing approaches for clinical trial matching can be divided into two categories depending on how matching is done. 

\paragraph{Structure-then-Match} Such systems first structure trial eligibility criteria by extracting key entities and relations for eligibility and then matching against structured patient information using manually crafted normalization rules to resolve superficial variations.
Prior systems typically require extensive human annotations~\citep{10.1136/amiajnl-2011-000321,10.1093/jamia/ocx019}, supervised learning models (for rule extraction)~\citep{bustos2018learning}, or a combination of supervised learning and rules (e.g., Criteria2Query~\citep{yuan2019criteria}).
Due to lexical variations and inadequate rule coverage, these systems often suffer from lower recall and generalizability. 
However, they can produce structured forms for eligibility criteria as intermediate results, which offer better interpretability and facilitate human-in-the-loop verification. %

\paragraph{End-to-End} These systems learn to encode patient and criteria for end-to-end matching via supervised learning from patient-trial or patient-criterion matching labels.  
For example, DeepEnroll~\citep{zhang2020deepenroll} jointly encodes patient records and trial eligibility criteria in the same embedding space, and then aligns them using attentive inference by learning from example patient-criteria matching data. COMPOSE~\citep{gao2020compose} represents the state-of-the-art embedding-based model that uses hierarchical memory network to encode concepts at different granularity and differentiates between inclusion and exclusion criteria. \citet{yuan2023llm} follows COMPOSE but uses \acp{LLM} to generate semantically equivalent criteria for data augmentation.

To the best of our knowledge, we are the first to explore using the emergent in-context learning capability of \acp{LLM}~\citep{NEURIPS2020_1457c0d6} for clinical trial matching.
Out of box and with no more than three examples, cutting-edge \acp{LLM}, such as GPT-4, can already structure trial eligibility criteria with reasonable performance, outperforming strong baselines from prior systems such as Criteria2Query. We also show preliminary results in applying \acp{LLM} to end-to-end clinical trial matching.

\section{Methods}

\begin{figure}[!ht]
    \centering
    \includegraphics[width=0.98\textwidth]{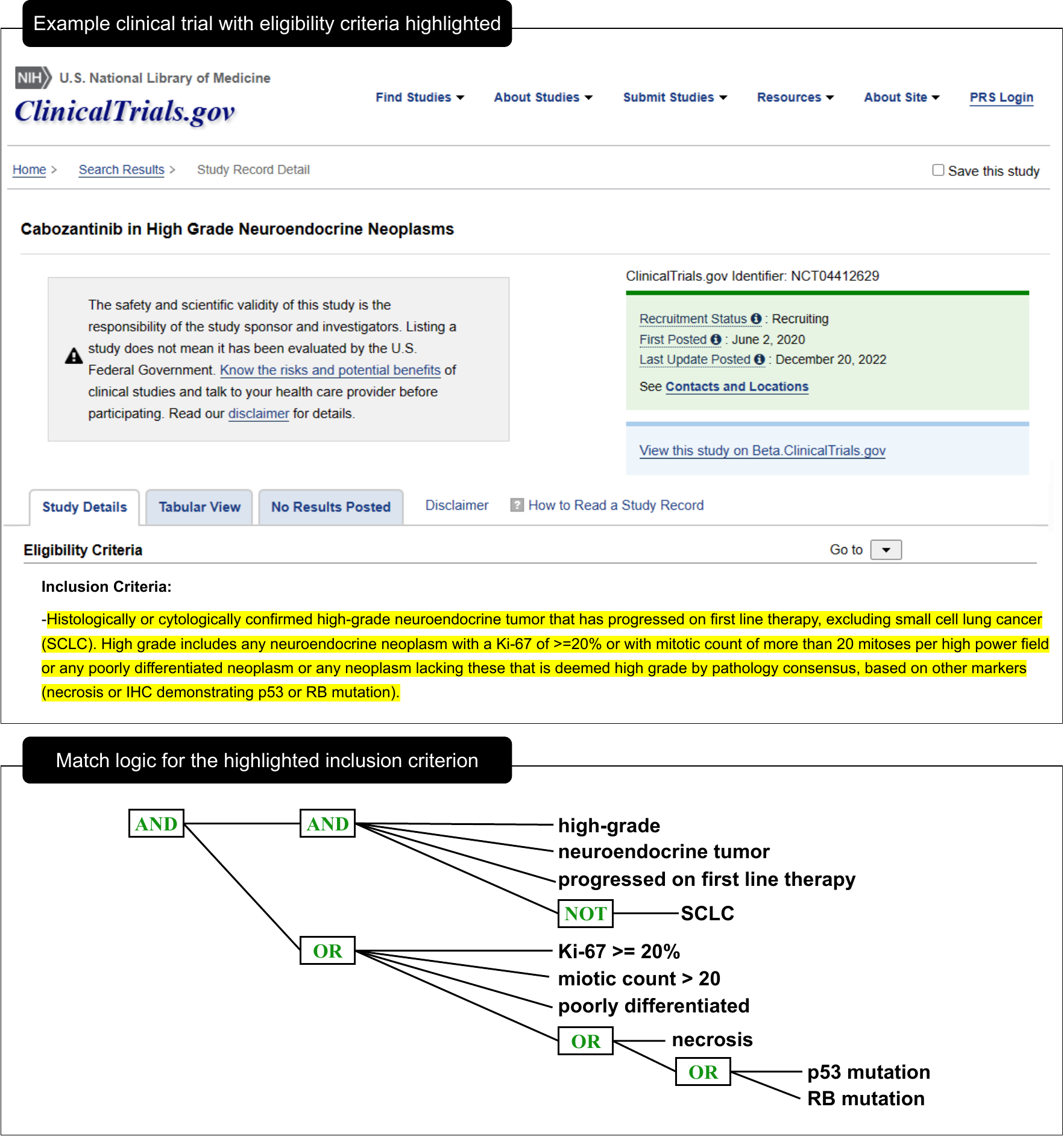}
    \caption{Example clinical trial (\href{https://clinicaltrials.gov/ct2/show/NCT04412629}{NCT04412629}) with eligibility criteria highlighted and the match logic for the inclusion criterion.}
    \label{fig:ec_example}
\end{figure}

In this section, we introduce the problem formulation for clinical trial matching and then describe the relevant systems and evaluation.

\subsection{Problem Formulation}

Given a clinical trial $T$ and a patient $P$, clinical trial matching determines if $P$ satisfies all eligibility criteria specified by $T$.
The specification $T_{\tt spec}$ generally comprises semi-structured text (see \Cref{fig:ec_example} ``Match logic'').
Similarly, the patient information from \acp{EMR} $P_{\tt EMR}$ also contains many free-text clinical documents, such as pathology reports, radiology reports, and encounter notes.
To determine the match, there are two approaches:
\begin{itemize}
    \item Structure-then-match: first structure both trial specification and patient information into common ontologies by extracting the relevant information, then execute the matching logic;
    \item End-to-end: learn to encode the trial and patient information directly and determine match by computing the similarity of embeddings.
\end{itemize}

DeepEnroll~\citep{zhang2020deepenroll} and COMPOSE~\citep{gao2020compose} are representative examples of the end-to-end approach. Learning an accurate encoding is particularly challenging for patient, given the vast amount of potentially relevant medical records. Consequently, prior work focuses on a more limited setting where the patient records are already structured. Still, such methods require large amount of training data of known patient-trial matches (or patient-criteria matches) and their generalizability to unseen disease areas is underexplored. It is also hard for human experts to interpret or verify such end-to-end results.

In this paper, we instead focus our study on ``structure-then-match". 
Let $T_{\tt structured}$ denote the structured representation of the trial eligibility criteria and $P_{\tt structured}$ denote that of a patient. Then clinical trial matching reduces to three subtasks:
\begin{itemize}
    \item Structuring trial eligibility criteria: $T_{\tt spec}\rightarrow T_{\tt structured}$;
    \item Structuring patient records: $P_{\tt EMR}\rightarrow P_{\tt structured}$;
    \item Matching: ${\tt Match}(T_{\tt structured}, P_{\tt structured})\rightarrow \{{\tt yes, no}\}$;
\end{itemize}

Matching is relatively straightforward, assuming that structuring has been done well, so we will focus on the structuring tasks.
At the time of this study, it is not yet possible to applying cutting-edge \acp{LLM}, such as GPT-4, to \ac{PHI} data in the environment available. Therefore, in the remainder of the paper, we focus on the structuring task for clinical trial eligibility criteria ($T_{\tt spec}\rightarrow T_{\tt structured}$). 

\subsection{Structuring Clinical Trial Eligibility Criteria}

Eligibility criteria comprise inclusion and exclusion criteria. There is an implicit \texttt{AND} logic over components in the inclusion criteria, and an implicit \texttt{NOT AND} logic over components in the exclusion criteria.
Additionally, there may be complex nested logic on combinations of patient attributes. 

For oncology, a salient example is the interplay between disease histology (fine-grained subtypes of disease states) and genomic biomarkers (e.g., genetic point mutation). See \Cref{fig:ec_example} for an example criterion. %
By contrast, standard demographic attributes (e.g., age, gender) are straightforward. We omit them for simplicity and focus our evaluation on disease histology and biomarkers.

\subsection{Systems}

Our main focus of the paper is to explore applying cutting-edge \acp{LLM}, such as GPT-3.5 and GPT-4, to the structuring tasks. 
For head-to-head comparison, we also consider a strong baseline using state-of-the-art biomedical entity extraction systems, prior state of the art Criteria2Query~\citep{yuan2019criteria}, and an expert system in test deployment at a large health network.

\paragraph{LLM} We use GPT-3.5~\citep{ouyang2022training} and GPT-4~\citep{openai2023gpt4} through Azure OpenAI Service,\footnote{\url{https://azure.microsoft.com/en-us/products/cognitive-services/openai-service/}} which allows us to interact with the models efficiently and in HIPAA-compliant way. Through the API, we submit prompts and receive generated responses, which we postprocess to the requisite structured format.

\paragraph{Biomedical Entity Extraction} In the absence of a general model such as LLMs, we can combine state-of-the-art biomedical entity extraction systems to assemble a strong baseline. Specifically, we consider SciSpaCy~\citep{neumann2019scispacy} for extracting disease states, GNormPlus~\citep{Wei2015GNormPlus} for extracting gene entities, and tmVar~\citep{Li2013TmVar} for extracting genetic variants.

\paragraph{Criteria2Query} \cite{yuan2019criteria} provide a standard baseline for structuring eligibility criteria. Criteria2Query combines supervised learning and rule-based heuristics and prior work shows state-of-the-art results in extracting disease states. We found that it can also extract biomarker information, although the capabilities are limited.

\paragraph{Expert System}\label{sec:expert-system} Some authors of this paper have previously developed an expert system for clinical trial matching, in collaboration with a large health network, where this system is currently in test deployment. This system comprises three key components: structuring trial eligibility criteria, matching against structured patient information, and a clinical trial matching application with a human-in-the-loop assisted clinical trial triaging user interface (UI). The structuring component encapsulates extensive heuristics for extracting and normalizing biomedical entities such as disease histology, genomic biomarkers, etc., as well as heuristics for processing semi-structured information (e.g., sections) and extracting matching directives (\texttt{AND}/\texttt{OR}/\texttt{NOT}). In internal evaluation, this system has demonstrated state-of-the-art performance and has been well received in test deployment by molecular tumor boards and trial coordinators. However, as common for rule-based approaches, this system has taken over 450 expert hours to develop and is specifically tailored for oncology (e.g., by focusing its structuring on disease histology and biomarkers). 
Exploring a more scalable approach thus represents an attractive direction for expanding to more fine-grained details and generalizing to other disease areas.

\subsection{Evaluation}

\begin{figure*}[!t]
    \centering
    \includegraphics[width=1.0\textwidth]{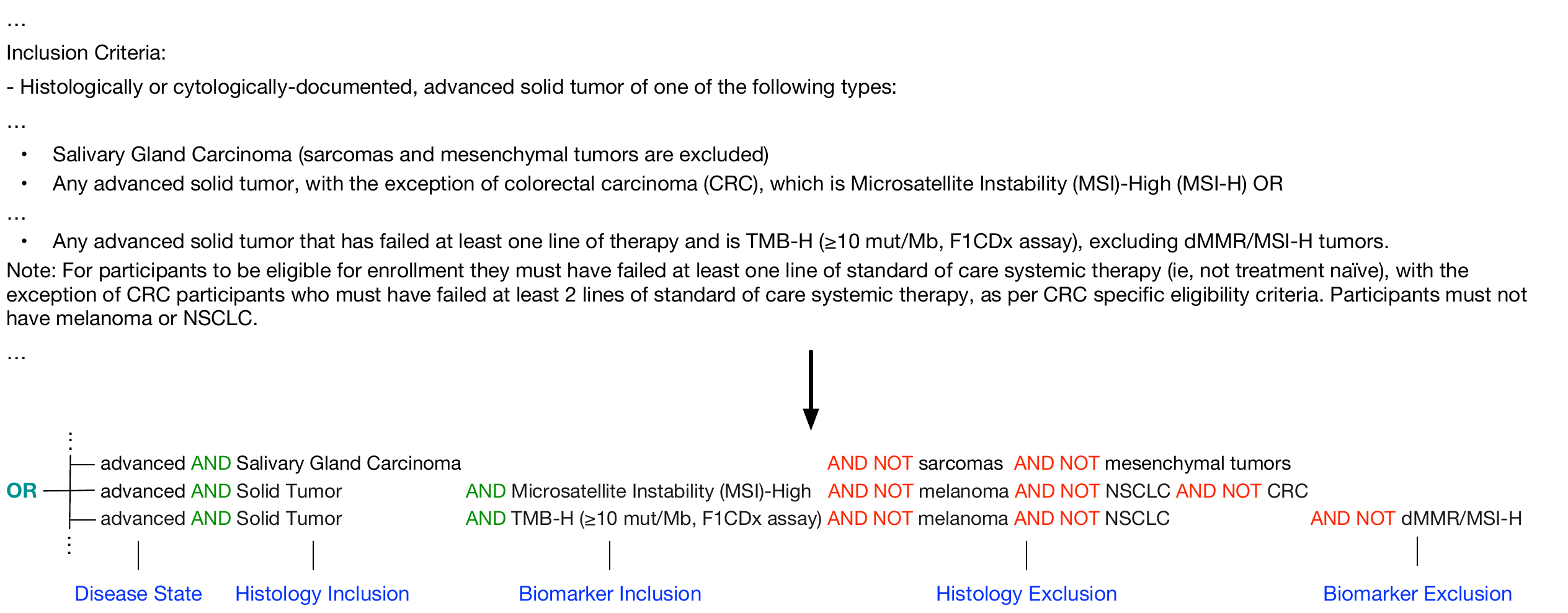}
    \caption{Example of disjunctive normal form (DNF). In DNF formulas must be a disjunction (\texttt{OR}) of conjunction (\texttt{AND}).}
    \label{fig:DNF}
\end{figure*}

\paragraph{Intrinsic Evaluation} For structuring trial eligibility criteria, we consider two settings. In the basic setting, we evaluate on entity extraction only, focusing on histology and biomarkers. Specifically, we compare system output against the gold entity set and report precision (i.e., positive predictive value), recall (i.e., sensitivity), and F1 (harmonic mean of precision and recall).
This is similar to the approach taken by prior work. In the more advanced setting, we evaluate on extraction of the complete matching logic. Essentially, we regard structuring as a semantic parsing problem, where the structuring output is a logic form as in \Cref{fig:ec_example} (see ``Match logic''). %
To facilitate evaluation, we normalize the logic form to \ac{DNF} (i.e., \texttt{OR} of \texttt{AND}s) and report precision/recall/F1 of the disjunctions. See \Cref{fig:DNF} for an example.

\paragraph{End-to-End Evaluation} Given the structured trial information, we can simulate an end-to-end evaluation by triangulating it with structured patient information via a common matching component to determine end-to-end match results. We can then compare these results against a gold match dataset.

\section{Experiments}

\subsection{Datasets}

\paragraph{Clinical Trial} We downloaded all the clinical trials in XML format from \href{https://clinicaltrials.gov/}{ClinicalTrials.gov}~\citep{clinicaltrials}. In this work, we focus on oncology trials with ``Treatment'' as the primary purpose and ``Interventional'' as the study type. We use a list of oncology-related keywords to filter out non-oncology trials. For each trial, we extract the following XML elements: \texttt{brief\_title}, \texttt{official\_title}, \texttt{brief\_summary}, \texttt{arm\_group}, and \texttt{criteria}. We truncate the criteria element to the first 40 lines and apply heuristics to remove lines that are not related to oncology-specific criteria (disease histology and biomarkers).

\paragraph{Structured Eligibility Criteria} To create the gold dataset for evaluating the structured criteria, three molecular pathologists at our collaborating health network manually structured 53 clinical trials as the test set. These 53 clinical trials were randomly selected from a larger set of trials that had significant enrollment at the health network and were more likely to contain complex criteria.
The output organizes the criteria into five categories: disease state, histology inclusion, biomarker inclusion, histology exclusion, and biomarker exclusion. The criteria logical expression is in disjunctive normal form (DNF), which consists of a disjunction of conjunctive clauses ($\tt OR$ of $\tt AND$s). See \Cref{fig:DNF}. %
Given the extracted histology terms and biomarkers, we further normalize them into the NCI Thesaurus ontology\footnote{\href{https://ncithesaurus.nci.nih.gov/ncitbrowser/}{https://ncithesaurus.nci.nih.gov/ncitbrowser/}} and HGVS nomenclature\footnote{\href{https://varnomen.hgvs.org/}{https://varnomen.hgvs.org/}}. %

\paragraph{Structured Patient Information} To facilitate simulated end-to-end matching evaluation, we leverage the component for structuring patient information in the expert system at our collaborating health network (see \Cref{sec:expert-system}). Briefly, this component system uses the Read OCR model in Azure Form Recognizer\footnote{\href{https://azure.microsoft.com/en-us/products/form-recognizer/}{https://azure.microsoft.com/en-us/products/form-recognizer/}} to convert any scanned images and PDF documents into free text, and then appends such text to digitized medical records. It then applies a series of state-of-the-art biomedical NLP models to extract relevant patient attributes from the medical records. For example, it uses self-supervised PubMedBERT models \citep{preston2023cellpattern} to extract the tumor site, histology, and staging information, as well as additional information extraction modules to extract other attributes such as health status, PD-L1 IHC result, and medications. Information available in structured EMRs is added directly, such as date of birth and gender. Other structured data fields such as lab tests and medications are normalized to the same units and NCI Thesaurus, respectively.

\paragraph{Historical Trial Enrollment Data} Given a known patient enrollment into a trial, we can treat it as a gold label and test if a clinical trial matching system can correctly flag it as a match. In this way, we can estimate the recall for end-to-end matching. 
For this purpose, we use a dataset containing 523 patient-trial enrollment pairs at our collaborating health system. This is a subset of all historical enrollment data, after filtering out confidential trials and patients who do not have sufficient structured information.

\begin{figure*}[!t]
    \centering
    \includegraphics[width=0.96\textwidth]{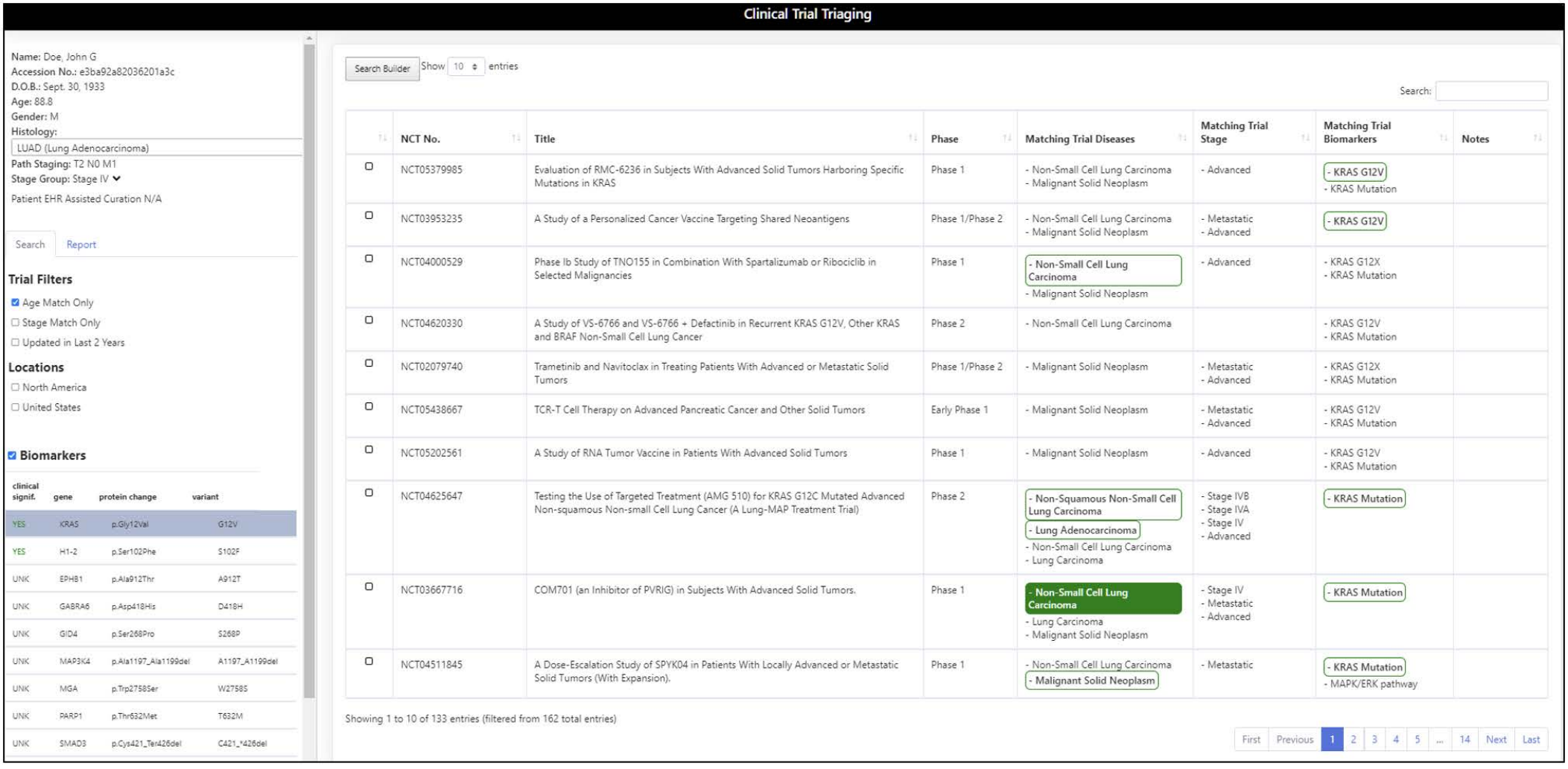}
    \caption{Example user interface for clinical trial matching. NOTE: The data used in this example are synthetic and do not contain PHI.}
    \label{fig:app}
\end{figure*}

\paragraph{Clinical Trial Matching System Feedback Data}
As mentioned in \Cref{sec:expert-system}, a clinical trial matching application is in test deployment at our collaborating health network (see \Cref{fig:app}), which allows molecular pathologists to inspect patient-trial match candidates and select the most suitable trials for follow-up. %
We used molecular pathologists' selection as implicit feedback and consider selected candidates as positive and non-selected ones as negative.
If no trials were selected for a patient, we skipped all candidates for this patient, as the molecular pathologists may not have inspected them yet. This yielded a dataset with 68,485 candidates with 84\% positive. %

\paragraph{Human Subjects, IRB, Data Security and Patient Privacy}
This work was performed under the auspices of an \ac{IRB}-approved research protocol 
(Providence protocol ID 2019000204)
and was conducted in compliance with Human Subjects research, clinical data management procedures, as well as cloud information security policies and controls. All study data involving \ac{PHI} were integrated, managed and analyzed exclusively and solely within our collaborating health network. All study personnel completed and were credentialed in training modules covering Human Subjects research, use of clinical data in research, and appropriate use of IT resources and \ac{IRB}-approved data assets.

\subsection{System Details: Applying LLMs to Structuring Eligibility Criteria}

To transform the trial XML into a structured representation, we employ a prompt template that guides GPT-4 (same for GPT-3.5) to extract and structure relevant criteria from each trial. Specifically, we focus on four types of criteria: trial cohort, disease state, %
tumor histology, and biomarkers. 
The prompt directs GPT-4 to output the structured representation in JSON format and provides instructions on how to handle unclear or missing information. In addition, the prompt may include few-shot example (input, output) pairs for in-context learning~\citep{NEURIPS2020_1457c0d6}.
Our prompt template can be found in \Cref{fig:prompt-template1,fig:prompt-template2,fig:prompt-template3,fig:prompt-template4} in Appendix.
For inference, we replace the placeholder \{\texttt{input\_trial}\} in the prompt template with the input trial text and prompt GPT-4 to output the structured representation in JSON format.
As shown in \Cref{fig:prompt-template4}, the output organizes the criteria into five categories: disease state, histology inclusion, biomarker inclusion, histology exclusion, and biomarker exclusion. The criteria logical expression is in disjunctive normal form (DNF). Our prompt instructs GPT-4 to assign a cohort name for each clause if possible.

\subsection{Results: Structuring Trial Eligibility Criteria}

\begin{table}[!ht]
    \centering
    \caption{Comparison of test results on structuring oncology trial eligibility criteria. Gold labels provided by three molecular pathologists at the collaborating health network. Evaluation on inclusion and exclusion criteria entity extraction only, with complex match logic ignored. GPT-3.5 can't do 3-shot here due to limited context size.}
    \scalebox{0.7}{
        \begin{tabular}{|l|r|r|r|r|r|r|}
    \hline
        & \multicolumn{3}{|c|}{Histology} & \multicolumn{3}{|c|}{Biomarker} \\
        & Precision & Recall & F1 & Precision & Recall & F1 \\ \hline
         GNormPlus & - & - & - & 6.8 & 19.6 & 10.2 \\ \hline
         SciSpaCy & 34.2 & 70.2 & 46.0 & 58.3 & 6.9 & 12.3 \\ \hline
         Criteria2Query & 29.6 & 40.2 & 32.8 & 68.3 & 27.5 & 39.2 \\ \hline
         GPT-3.5 (zero-shot) & 35.1 & 31.6 & 34.2 & 61.2 & 29.4 & 39.7 \\ \hline
         GPT-4 (zero-shot) & 62.1 & 69.0 & {\bf 65.4} & 75.3 & 59.8 & 66.7 \\ \hline
         GPT-4 (3-shot) & 57.8 & 73.7 & 64.8 & 72.5 & 72.5 & {\bf 72.5} \\ \hline
    \end{tabular}
    }
    \label{tab:entity-extraction} 
\end{table}

\begin{table}[!ht]
    \centering
    \caption{Comparison of test results on structuring oncology trial eligibility criteria, evaluated on complete match logic in DNF form.}
    \scalebox{0.7}{
    \begin{tabular}{|l|l|l|l|l|l|l|l|l|l|}
    \hline
        & \multicolumn{3}{|c|}{Histology} & \multicolumn{3}{|c|}{Biomarker} & \multicolumn{3}{|c|}{Histology+Biomarker} \\
        & Precision & Recall & F1 & Precision & Recall & F1 & Precision & Recall & F1 \\ \hline
        SciSpaCy & 9.2 & 16.8 & 11.8 & 36.4 & 5.3 & 9.3 & 7.4 & 10.8 & 8.8 \\ \hline
         Criteria2Query & 23.2 & 34.8 & 27.8 & 22.0 & 12.0 & 15.5 & 9.6 & 17.4 & 12.4 \\ \hline
         GPT-3.5 (zero-shot) & 36.0 & 20.6 & 26.2 & 48.3 & 18.7 & 26.9 & 30.7 & 14.6 & 19.7 \\ \hline
         GPT-4 (zero-shot) & 43.0 & 41.3 & 42.1 & 31.0 & 17.3 & 22.2 & 27.8 & 21.1 & 24.0 \\ \hline
         GPT-4 (3-shot) & 42.7 & 54.8 & {\bf 48.0} & 39.4 & 37.3 & {\bf 38.4} & 27.3 & 32.4 & {\bf 29.6} \\ \hline
    \end{tabular}
    }
    \label{tab:complete-logic} 
\end{table}

\Cref{tab:entity-extraction} shows the test results on structuring oncology trial eligibility criteria in the basic setting, where we evaluate on inclusion and exclusion criteria entity extraction only and ignore complex match logic. We can't get GPT-3.5 to work in the 3-shot setting as adding the examples will render the prompt to exceed the context limit. Standard biomedical entity extraction tools such as GNormPlus~\citep{Wei2015GNormPlus}, tmVar~\citep{Li2013TmVar}, and SciSpaCy~\citep{neumann2019scispacy} did not perform very well for one or both of histology and biomarkers. Criteria2Query~\citep{yuan2019criteria} also delivered subpar performance on oncology trial entities, revealing limitation in its generalizability. Remarkably, out of box, GPT-4 delivers strong performance in extracting both histology and biomarker information, despite requiring no extensive customization for this domain and task, outperforming Criteria2Query by over 30 absolute points in F1. 
In-context learning helped substantially for biomarker extraction, but didn't matter much for histology.
GPT-4 also clearly outperforms GPT-3.5, which performs on par with Criteria2Query in the zero-shot setting.

\Cref{tab:complete-logic} shows the test results on structuring oncology trial eligibility criteria in the full setting, where we evaluate on the complete match logic in DNF form. Note that this is a very stringent evaluation as each conjunction ($\tt AND$s) needs to contain the exactly combination of atomic attributes to be considered correct. When there are complex nested logical forms, the number of conjunctions may become very large.
Still, this evaluation offers a way to objectively compare system's capability in extracting complex match logic.
Here, GPT-4 similarly dominates all other systems. Moreover, GPT-4 (3-shot) outperforms GPT-4 (zero-shot) in all scenarios, indicating that in this more challenging setting, in-context learning indeed plays a positive role.

\begin{table}[!ht]
    \centering
    \caption{Comparison of recall results on historical patient-trial enrollment data. The expert system takes extensive expert manual effort to develop and has been designed to favor recall and ignore exclusion criteria, so it is expected to have high recall. The publicly available Criteria2Query instance runs slowly and we can only get 157 pairs evaluated in time (out of 466 total).}
    \begin{tabular}{|l|l|}
    \hline
        & Recall \\ \hline
        SciSpaCy & 50.0 \\ \hline
        Criteria2Query* & 37.0 \\ \hline
        GPT-3.5 (zero-shot) & 23.4 \\ \hline
        GPT-4 (zero-shot) & 64.2 \\ \hline
        GPT-4 (3-shot) & 76.8 \\ \hline
        Expert System &  93.6 \\ \hline
    \end{tabular}
    \label{tab:legacy-enrollment}
\end{table}

\begin{table}[!ht]
    \centering
    \caption{Comparison of test results using feedback data from the clinical trial matching application in test deployment at our collaborating health network. The expert system is expected to perform well here, given that the users mainly evaluated its proposed candidates.}
        \begin{tabular}{|l|r|r|r|}
    \hline
        & Precision & Recall & F1 \\ \hline
         GPT-3.5 (zero-shot) & 88.5 & 11.6 & 20.6 \\ \hline
         GPT-4 (zero-shot) & 86.7 & 46.8 & 60.8 \\ \hline
         GPT-4 (3-shot) & 87.6 & 67.3 & 76.1 \\ \hline
         Expert System & 84.3 & 97.4 & 90.4 \\ \hline
    \end{tabular}
    \label{tab:CTM-app}
\end{table}

\subsection{Results: End-to-End Clinical Trial Matching}

Given structured trial eligibility criteria and structured patient information, matching can be done by evaluating for each criterion component, where the patient information is compatible (thus eligible). This check is more elaborate than it may appear at first, since a criterion may specify a set of values (e.g., EGFR mutations without specifying which ones) or a more abstract concept than that in the patient information. 
Therefore, matching relies on a hierarchical representation of medical terms to check if a criterion entity subsumes the corresponding patient attribute. 
For histology matching, we convert both clinical trial criteria entities and patient tumor site/histology into OncoTree codes~\citep{doi:10.1200/CCI.20.00108}, which are a standard hierarchical representation for cancer types. 

For biomarker matching, we use a hierarchy that captures biomarkers at different levels of specificity: amino acid level, exon level, chromosomal level, gene level, or pathway level. A patient's tumor sequencing results typically specify a specific amino acid or DNA coding variant, whereas a clinical trial may specify a biomarker criterion at any of these levels. Matching needs to account for this hierarchy and the implied subsumption relations. %

\Cref{tab:legacy-enrollment} shows the recall results for the historical patient-trial enrollment dataset. As expected, the expert system performs very well, given that it has been developed with extensive manual effort for the oncology domain. Remarkably, out of box, GPT-4 can already performs competitively, recovering 76.8\% of gold patient-trial pairs.

\Cref{tab:CTM-app} shows the test results using the feedback data from the test deployment of the expert system. Again, the expert system is expected to perform well here. E.g., the recall is expected to be close to 100\%, given that the users mainly evaluated candidates proposed by the expert system and only occasionally added a trial they found by themselves. Out of box, GPT-4 already attains higher precision than the expert system. As observed in the historical enrollment evaluation, a major growth area lies in boosting the recall but GPT-4 already performs quite competitively overall.

We also explored applying GPT-4 directly to conduct end-to-end matching. Due to the context limit, we can't fit the entire medical record for most patients into the prompt. Therefore, it is only feasible to consider direct matching against patient information that has already been structured, and currently it is only feasible for the zero-shot setting.
Due to current restriction on applying Azure OpenAI services to PHI, we only test this using a de-id example adapted from the public TCGA dataset. The results can be found in \Cref{fig:text_match_ex_p1,fig:text_match_ex_p2,fig:text_match_ex_p3}. Remarkably, GPT-4 provides a cogent narrative of its matching decision, supported by detailed analysis of individual criteria. It is difficult to draw a generalizable conclusion from such an anecdotal example, but in combination with other experimental results, it certainly illustrates the potential in harnessing the general cognitive capabilities of large language models to scale clinical trial matching.

\section{Discussion} 
In this work, we present the first systematic study on using large language models (LLMs) to scale clinical trial matching, using oncology as a case study to ground our exploration. Our findings foremost suggest that out-of-the-box, \acp{LLM} such as GPT-4 can already handle complex eligibility criteria of clinical trials and extract complex matching logic. In this regard, \ac{LLM} provide a strong, practical baseline. And its general nature bodes well for the potential to apply to other disease areas.

Our study also reveals several key growth areas for enhancing \acp{LLM} in end-to-end clinical trial matching, such as addressing context length limitation and accuracy issues, especially in extracting and structuring patient data from \acp{EMR}. In future work, we plan to implement more sophisticated prompt engineering techniques and \ac{LLM} fine-tuning.

While there are implicit limitations in the use of logical expressions, 
our system enables a triaging pipeline that can reduce the clinical trial candidates to a small number. This enables human-in-the-loop participation wherein a human can manually go through the small number of trials to verify all the criteria are met. Human-in-the-loop may be preferable given the safety-critical nature of healthcare. Nevertheless, we experimented with further accelerating this process by providing GPT-4 structured patient information, and asking GPT-4 to output all the matching and non-matching conditions. This may not be efficient to do for all possible patient-trial pairs but we can reserve this as a more expensive but higher-quality reranking step on a small number of candidates produced by a more efficient but less powerful system.

\paragraph{Limitations}

Clinical trial matching by structuring all criteria into a logical form (i.e. ``structure-then-match'') carries implicit limitations. In particular, it is not always possible to map an extracted criteria into an existing concept or ontology; and indeed, we found quite a few examples in our study where this was not possible. There are also various subtleties in the criteria language that are difficult to capture completely into a logical formula. Due to current restriction on applying GPT-4 to identifiable patient records, we were not able to explore LLMs in structuring patient information.

\bibliography{anthology,references}
\newpage
\appendix
\section{Prompt Templates}

\begin{figure}[!ht]
\begin{AIbox}{Prompt Template for Clinical Trial Structuring}
{\bf (System Message)} I am an intelligent and thorough agent designed is to extract the given clinical trial criteria input into a valid JSON logic structure. My response should be written in the language of JSON and should accurately capture the meaning of the input. Please note that my response should accurately reflect the logical connectives of the criteria (and, or, not). I will try my best attempt to normalize the logic formula into disjunctive normal form. In disjunctive normal form, each clause connected by a conjunction, or 'and', must be either a literal or contain a conjunction, or 'and' operator.  Please keep in mind that my response should be flexible enough to allow for various relevant and creative solutions. I should also focus on providing an accurate and well-structured solution that can be easily understood by others. I should be sure to specify solid tumor and tumor staging when applicable. I only extract the most important tumor biomarkers, tumor histology or disease, and tumor staging information. I ignore criteria regarding life expectancy, written consent, pregnancy, contraception and adequate bone marrow, liver, and renal function. I ignore the medication treatment and dosages for trial arms. I only extract about the inclusion and exclusion criteria for the trial. I must be thorough and include all criteria related to biomarkers, histologies, disease state, and staging.

---

{\bf (Instructions)} I will use the following output format:\\

[\{``cohort": (String) Cohort Name, ``disease\_state":(String) Disease State Criteria, ``histology\_inclusion": (String) Histology Inclusion Criteria,  ``biomarker\_inclusion": (List) Biomarker Inclusion Criteria, ``histology\_exclusion": (List) Histology Exclusion Criteria, ``biomarker\_exclusion": (List) Biomarker Exclusion Criteria\}, ...] \\

This is a representation of the logical disjunctive normal form where each conjunctive (AND) clause is represented as a JSON and the outer list is a disjunction (logical OR) of those clauses or JSONs. I will try to separate the biomarkers and histologies.\\ 

Below are descriptions of those keys. Only the following keys are allowed. DO NOT include any other criteria keys:

cohort: a string that represents the name of the cohort for these criteria

disease\_state: a string that describes the disease state (e.g, staging (stage I, II, IIIA, IV), refractory, relapsed, advanced, metastatic, etc.)

histology\_inclusion: a string that represents the histology criteria that if satisfied, contributes to the expression evaluating to True. There should only be one histology per JSON item.
\end{AIbox}
\caption{GPT-4 prompt template for clinical trial structuring.}\label{fig:prompt-template1}
\end{figure}

\begin{figure}[!t]
\begin{AIbox}{Prompt Template for Clinical Trial Structuring (cont'd)}
biomarker\_inclusion: a list of biomarker inclusion criteria that if all elemnts of the list are satisfied, contributes to the expression evaluating to True.

histology\_exclusion: a list of histology exclusion criteria if any elements of the list are satisfied, contributes to the expression evaluating to False.

biomarker\_exclusion: a list of biomarker exclusion criteria that if any elements of the list are satisfied, contributes to the expression evaluating to False.\\

To match one of the clauses or JSON items in the list, it means all field values in each item are satisfied ("histology\_inclusion" and "biomarker\_inclusion" and not any("histology\_exclusion") and not any("biomarker\_exclusion")). Each clause/JSON should have at least a value for histology\_inclusion. If all the values for a JSON are empty or null, do you include in the list. I do not care about criteria that do not have at least a value histology\_inclusion. If a field doesn't have any value, set the value to "" or []. Any of the JSON in the list matching means a patient is eligible for the trial. For criteria specific to a cohort, add a new item to the list. Skip biomarker or histology criteria that are inside if-then conditions unless the if condition is a cohort type. Skip prior treatment type criteria. If the extracted criteria only contains an abbreviation, I will find the long form of that abbreviation from the rest of the document and include that in the field value.\\

I recognize the following as biomarkers and will include them in the biomarker\_inclusion or biomarker\_exclusion fields if they are mentioned as criteria as defined above: gene mutations, gene expressions, TMB (tumor molecular burden, TMB-H or TMB-L), MSI (microsatellite instability, MSI-H, MSI-L, MSS, microsatellite stable) status, gene pathway alterations such as dMMR (deficient Mismatch Repair Pathway) or pMMR (proficient Mismatch Repair), PD-L1 expression levels determined by Combined Positive Score (CPS) or tumor proportion score (TPS), positive or negative status for breast cancer biomarkers such as HR, ER, PR and HER2.\\

When a criteria specifies a histology and all the subtypes of the histology, I will include all those subtypes as additional clauses (JSON) in the list of disjunctions.\\

When a trial exclusion includes primary brain (Central Nervous System, CNS) tumor (not just brain metastasis), also include brain / CNS tumor mention in the histology\_exclusion list.\\

Do not include criteria about prior therapies or prior treatment. They are not considered biomarkers criteria and should not be included in the inclusion or exclusion criteria. Do not include any expression prior therapies, treatments, or therapies.
\end{AIbox}
\caption{GPT-4 prompt template for clinical trial structuring (cont'd).}\label{fig:prompt-template2}
\end{figure}

\begin{figure}[!t]
\begin{AIbox}{Prompt Template for Clinical Trial Structuring (cont'd)}
If the inclusion criteria is for either of multiple biomarkers (and/or), list each of those biomarkers in a separate clause or JSON item because either of them being satisfied should contribute to the entire expression being True. I want to have the least restrictive accurate matching criteria output. Only list multiple biomarkers in biomarker\_inclusion list for one clause JSON item if the trial absolutely require all of those biomarkers.

Do not include mentioned biomarkers if the presence or absense of the biomarker does not affect eligibility. And do not include biomarkers in the output that if the biomarker is present, then additional criteria is needed. I only extract if the criteria itself determines eligibility.\\
 
When there is a list of biomarker criteria in the same sentence and it's not clear whether the biomarkers have an AND or OR relationship, assume it's an OR criteria relationship between the biomarkers, which means assign each biomarker to a new clause. If the biomarkers are on separate main bullet points in criteria section, assume those are AND criteria relationship and all the biomarkers should be in the same clause.\\

Do not include cohort or arms that closed.\\

For multiple exclusion biomarker or histology criteria, those should be present in every clause JSON item.

DO NOT repeat the same JSON multiple times. Be succinct.

---

{\bf (Demonstrations)} 

Input:
\lstset{language=XML,basicstyle=\ttfamily\scriptsize,breaklines=true}
\begin{lstlisting}
<brief_title>A Phase 1/2 Study of DCC-3116 in Patients With MAPK Pathway Mutant Solid Tumors</brief_title>
<official_title>A Phase 1/2, First-in-Human Study of DCC-3116 as Monotherapy and in Combination With RAS/MAPK Pathway Inhibitors in Patients With Advanced or Metastatic Solid Tumors With RAS/MAPK Pathway Mutations</official_title>
<brief_summary>This is a Phase 1/2, multicenter, open label, first in human (FIH) study of DCC-3116 as monotherapy, and in combination with trametinib, binimetinib, or sotorasib in patients with advanced or metastatic solid tumors with RAS/MAPK pathway mutation. The study consists of 2 parts, a dose-escalation phase, and an expansion phase.</brief_summary>
<condition>Pancreatic Ductal Adenocarcinoma</condition>
<condition>Non-Small Cell Lung Cancer</condition>
<condition>Colorectal Cancer</condition>
<condition>Advanced Solid Tumor</condition>
<condition>Metastatic Solid Tumor</condition>
\end{lstlisting}
\end{AIbox}
\caption{GPT-4 prompt template for clinical trial structuring (cont'd).}\label{fig:prompt-template3}
\end{figure}

\begin{figure}[!t]
\begin{AIbox}{Prompt Template for Clinical Trial Structuring (cont'd)}
\lstset{language=XML,basicstyle=\ttfamily\scriptsize,breaklines=true}
\begin{lstlisting}
<arm_group>
<arm_group_label>Dose Escalation (Part 1, Cohort A Monotherapy)</arm_group_label>
<arm_group_type>Experimental</arm_group_type>
<description>DCC-3116 tablets in escalating dose cohorts given orally twice daily (BID) in 28-day cycles as monotherapy (single agent). If no DLT in 3 participants or 1 DLT/6 participants is observed, dose escalation may continue to the next planned dose cohort.</description>
</arm_group>
\end{lstlisting}
{\ttfamily\scriptsize \textcolor{gray}{(... more arm groups are omitted for space saving)}}
\lstset{language=XML,basicstyle=\ttfamily\scriptsize,breaklines=true}
\begin{lstlisting}
<criteria>
Inclusion Criteria:
  1. Male or female participants >=18 years of age
  2. Dose Escalation Phase (Part 1):
    1. Participants must have a pathologically confirmed diagnosis of an advanced or metastatic solid tumor with a documented RAS, NF1, or RAF mutations. A molecular pathology report documenting mutational status of RAS, NF1, or RAF must be available.
    2. Progressed despite standard therapies, and received at least 1 prior line of anticancer therapy.
       -  Participants with a documented mutation in BRAF V600E or V600K must have received approved treatments known to provide clinical benefit prior to study entry.
\end{lstlisting}
{\ttfamily\scriptsize \textcolor{gray}{(... more criteria are omitted for space saving)}

</criteria>
}\\

Eligibility Criteria Output:
\lstset{language=Python,basicstyle=\ttfamily\scriptsize,breaklines=true,showstringspaces=false}
\begin{lstlisting}
[
    {"cohort": "Dose Escalation Phase (Part 1)", disease_state": "advanced or metastatic", "histology_inclusion": "Solid Tumor",  "biomarker_inclusion": ["RAS mutation"], "histology_exclusion": [], "biomarker_exclusion": []},
    {"cohort": "Dose Escalation Phase (Part 1)", "disease_state": "advanced or metastatic", "histology_inclusion": "Solid Tumor",  "biomarker_inclusion": ["NF1 mutation"], "histology_exclusion": [], "biomarker_exclusion": []}, {"cohort": "Dose Escalation Phase (Part 1)", "disease_state": "advanced or metastatic", "histology_inclusion": "Solid Tumor",  "biomarker_inclusion": ["RAF mutation"], "histology_exclusion": [], "biomarker_exclusion": []},
\end{lstlisting}
{\ttfamily\scriptsize \textcolor{gray}{(... more cohorts are omitted for space saving)}

]}\\

{\bf (User Message)}
Input:

\{\texttt{input\_trial}\}\\

Eligibility Criteria Output:
\end{AIbox}
\caption{GPT-4 prompt template for clinical trial structuring (cont'd).}\label{fig:prompt-template4}
\end{figure}

\begin{figure}[!ht]
\begin{AIbox}{Prompt Template for Direct Trial-Patient Matching}
{\bf (System Message)} As a helpful agent, my task is to analyze both the structured data and free-text pathology report of a patient and determine
if they meet the clinical trial criteria text for eligibility. Please provide a clear and concise response that lists all the
reasons why I believe the patient is eligible or not eligible. My analysis should take into account various factors such as
age, medical history, current medications, histology, staging, biomarkers and any other relevant information provided in the
report and structured data. I should also be able to handle complex scenarios with multiple variables and provide accurate
reasoning behind my determination. My response should be flexible enough to allow for various relevant and creative reasons
why the patient may or may not be eligible. My goal is to provide an informative and detailed analysis that assists healthcare
professionals in making informed decisions regarding patient eligibility.\\

Clinical Trial Study Design Detail:

\lstset{language=XML,basicstyle=\ttfamily\scriptsize,breaklines=true}
\begin{lstlisting}
<brief_title>Study of Zotiraciclib for Recurrent High-Grade Gliomas With Isocitrate Dehydrogenase 1 or 2 (IDH1 or IDH2)
Mutations</brief_title>
...
<criteria>
- INCLUSION CRITERIA:
- Participants must have diffuse glioma, WHO grades 2-4, histologically confirmed by Laboratory of Pathology, NCI
- IDH1 or IDH2 mutation status confirmed by TSO500 performed in LP, NCI
- Participants must have received prior treatment (e.g., radiation, conventional chemotherapy) prior to disease
progression
- Participants must have recurrent disease, proven histologically or by imaging studies
- Participants who have undergone prior surgical resection are eligible for enrollment to cohorts 1-4.
- Age &gt;=18 years
- Karnofsky &gt;=70%
- Participants must have recovered from the adverse effects of prior therapy to grade 2 or less
EXCLUSION CRITERIA:
More than one prior disease relapse (WHO grade 3-4) or more than two prior disease relapses (WHO grade 2)
- Prior therapy with:
- bevacizumab for tumor treatment. Note: participants who received bevacizumab for symptom management,
including but not limited to cerebral edema, or pseudo progression can be enrolled
- Prolonged QTc &gt;470ms as calculated by Fridericia s correction formula on screening electrocardiogram (ECG)
...
</criteria>
\end{lstlisting}
---
\end{AIbox}
\caption{Directly asking GPT-4 to match clinical trial criteria and patient record. The patient record is de-ID clinical data from public TCGA dataset.}\label{fig:text_match_ex_p1}
\end{figure}

\begin{figure}[!ht]
\begin{AIbox}{Prompt Template for Direct Trial-Patient Matching (cont'd)}
Patient Metadata:
\lstset{basicstyle=\ttfamily\scriptsize,breaklines=true}
\begin{lstlisting}
Male,
29 years old,
Glioma (Anaplastic Astrocytoma),
LIVING (63 months),
Recurred/Progressed (53 months)
Samples: 1
Attribute Value
....
International Classification of Diseases for Oncology, Third Edition ICD-O-3 Histology Code 9401/3
International Classification of Diseases for Oncology, Third Edition ICD-O-3 Site Code C71.0
Karnofsky Performance Score 90
...
Neoplasm Histologic Grade G3
Neoplasm Histologic Type Name Astrocytoma
...
Performance Status 0
Performance Status Assessment Timepoint Category Pre-Adjuvant Therapy
...
Related symptom first present Seizures
Seizure History YES
Sex Male
...
\end{lstlisting}
Patient Mutations:
\lstset{basicstyle=\ttfamily\scriptsize,breaklines=true}
\begin{lstlisting}
Missense IDH1 R132H
\end{lstlisting}
Pathology Metadata:
\lstset{basicstyle=\ttfamily\scriptsize,breaklines=true}
\begin{lstlisting}
Oncotree Code AASTR
...
\end{lstlisting}
\lstset{basicstyle=\ttfamily\scriptsize,breaklines=true}
Patient Pathology Report:
\begin{lstlisting}
SPECIMEN(S) SUBMITTED/ PROCEDURES ORDERED:
CLINICAL HISTORY: male, here for left side craniotomy for tumor.
GROSS DESCRIPTION: Received the following specimens in the Department of Pathology, labeled with
the patient's name and
A. Brain tumor
B. Brain tumor
...
\end{lstlisting}
DIAGNOSIS:
\lstset{basicstyle=\ttfamily\scriptsize,breaklines=true}
\begin{lstlisting}
A. "Brain tumor":
Anaplastic astrocytoma (WHO Grade III) (see comment)
B. "Brain tumor":
Anaplastic astrocytoma (WHO Grade III)
...
\end{lstlisting}
---
\end{AIbox}
\caption{Directly asking GPT-4 to match clinical trial criteria and patient record. The patient record is de-ID clinical data from public TCGA dataset (cont'd).}\label{fig:text_match_ex_p2}
\end{figure}

\begin{figure}[!ht]
\begin{AIbox}{Prompt Template for Direct Trial-Patient Matching (cont'd)}
Based on the provided information, the patient appears to be eligible for the clinical trial. Here are the reasons for
eligibility:

1. Age: The patient is 29 years old, which meets the age requirement of being 18 years or older.

2. Diagnosis: The patient has a diagnosis of Anaplastic Astrocytoma, which is a high-grade glioma (WHO Grade III). The
clinical trial is for recurrent high-grade gliomas with IDH1 or IDH2 mutations.

3. IDH1 Mutation: The patient has an IDH1 R132H mutation, which confirms the presence of an IDH1 mutation as required by the
clinical trial.

4. Recurrent Disease: The patient's disease has recurred/progressed after 53 months, which meets the requirement of having
recurrent disease.

5. Karnofsky Performance Score: The patient has a Karnofsky Performance Score of 90, which is greater than or equal to the
required 70\%.

6. Prior Treatment: The patient has received prior treatment (radiation, conventional chemotherapy) before disease
progression, as required by the clinical trial.

7. Recovery from Adverse Effects: Although not explicitly mentioned, there is no indication that the patient has not recovered
from the adverse effects of prior therapy to grade 2 or less.

However, there are some limitations in the provided information:

1. IDH2 Mutation: The IDH2 mutation status is not provided. However, since the patient has an IDH1 mutation, this is not a
concern for eligibility.

2. Prior Therapy Exclusions: The patient's prior therapy with bevacizumab is not mentioned. However, there is no indication
that the patient has received bevacizumab for tumor treatment, which would exclude them from the trial.

3. QTc Interval: The patient's QTc interval is not provided. However, there is no indication that the patient has a prolonged
QTc $>$470ms, which would exclude them from the trial.

4. Other Exclusion Criteria: There is no information provided about prior invasive malignancies, history of allergic reactions
to compounds similar to zotiraciclib, or uncontrolled intercurrent illness or social situations. However, there is no
indication that the patient meets any of these exclusion criteria.

In conclusion, based on the available information, the patient appears to be eligible for the clinical trial.
\end{AIbox}
\caption{Directly asking GPT-4 to match clinical trial criteria and patient record. The patient record is de-ID clinical data from public TCGA dataset (cont'd).}\label{fig:text_match_ex_p3}
\end{figure}

\end{sloppypar}
\end{document}